\documentclass[10pt, a4paper]{article}
\usepackage{lrec}
\usepackage{graphicx}
\usepackage{tabularx}
\usepackage{soul}
\usepackage{multirow}
\usepackage{epstopdf}
\usepackage[utf8]{inputenc}
\usepackage[T1]{fontenc}

\usepackage{hyperref}
\usepackage{xstring}

\usepackage{tikz}
\usetikzlibrary{trees, matrix}
\usepackage{forest}

\title{Kvistur 2.0: a BiLSTM Compound Splitter for Icelandic}

\name{Jón Friðrik Daðason\textsuperscript{\normalfont 1}, David Erik Mollberg\textsuperscript{\normalfont 1}, Hrafn Loftsson\textsuperscript{\normalfont 1}, Kristín Bjarnadóttir\textsuperscript{\normalfont 2}}

\address{\textsuperscript{1}Department of Computer Science, \textsuperscript{2}The Árni Magnússon Institute for Icelandic Studies \\
         \textsuperscript{1}Reykjavik University, \textsuperscript{2}University of Iceland \\
         \textsuperscript{1}\{jond19, de14, hrafn\}@ru.is, \textsuperscript{2}kristinb@hi.is\\}

\abstract{
In this paper, we present a character-based BiLSTM model for splitting Icelandic compound words, and show how varying amounts of training data affects the performance of the model. Compounding is highly productive in Icelandic, and new compounds are constantly being created. This results in a large number of out-of-vocabulary (OOV) words, negatively impacting the performance of many NLP tools. Our model is trained on a dataset of 2.9 million unique word forms and their constituent structures from the Database of Icelandic Morphology. The model learns how to split compound words into two parts and can be used to derive the constituent structure of any word form. Knowing the constituent structure of a word form makes it possible to generate the optimal split for a given task, e.g., a full split for subword tokenization, or, in the case of part-of-speech tagging, splitting an OOV word until the largest known morphological head is found. The model outperforms other previously published methods when evaluated on a corpus of manually split word forms. This method has been integrated into Kvistur, an Icelandic compound word analyzer.\\ \newline 
}

\begin{document}

\maketitleabstract

\section{Introduction}
Compounds are extremely common in Icelandic, accounting for over 88\% of all words in the \textit{Database of Icelandic Morphology} (DIM) \cite{Bjarnadottir_2017,Bjarnadottir_2019}. As compounding is so productive, new compounds frequently occur as out-of-vocabulary (OOV) words, which may adversely affect the performance of NLP tools. Furthermore, Icelandic is a morphologically rich language with a complex inflectional system. There are 16 inflectional categories (i.e., word forms with unique part-of-speech (PoS) tags) for nouns, for adjectives 120, and for verbs 122, excluding impersonal constructions. The average number of inflectional forms per headword  in DIM is 21.7. Included in this average are all uninflected words as well inflectional variants, i.e., dual word forms with the same PoS tag.

\begin{figure}[!h]
\begin{center}
\begin{forest}
  for tree={s sep=10pt, align=center}
  [heildarraforkuþörf\\‘total electric energy requirement’
    [heildar\\‘total’]
    [raforkuþörf\\‘electric energy requirement’
      [raforku\\‘electric energy’
        [raf\\‘electric’]
        [orku\\‘energy’]]
      [þörf\\‘need’]]]
\end{forest}
\caption{The constituent structure of an Icelandic compound word consisting of four constituent parts.}
\label{fig:tree}
\end{center}
\end{figure}
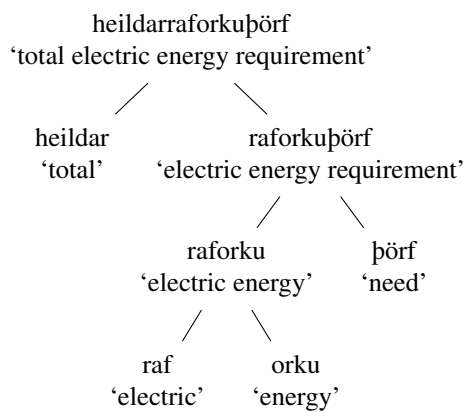

Compounds are formed by combining two words, which may be compounds themselves. The former word is known as a modifier and the second as a head, assuming binary branching \cite{Bjarnadottir_2005}. Theoretically, there is no limit to how many constituents a compound can be composed of, although very long words such as \textit{uppáhaldseldhúsinnréttingaverslunin} ‘the favorite kitchen furniture store’ (containing 7 constituent parts) are rare. The constituent structure of a compound word can be represented by a full binary tree, as shown in Figure \ref{fig:tree}.

Compound splitting, or decompounding, is the process of breaking compound words into their constituent parts. This can significantly reduce the number of OOV words for languages where compounding is productive. Compound splitting has been shown to be effective for a variety of tasks, such as machine translation \cite{Brown_2002,Koehn_2003}, speech recognition \cite{Adda-Decker_2000} and information retrieval \cite{Braschler_2002}.

In this paper, we present a character-based bidirectional long short-term memory (BiLSTM) model for splitting Icelandic compound words, and evaluate its performance for varying amounts of training data. Our model is trained on a dataset of 2.9 million unique word forms and their constituent structures from DIM. The model learns how to split compound words into two parts and can be used to derive the constituent structure of any word form. The model outperforms other previously published methods when evaluated on a corpus of manually split word forms. Our method has been integrated into Kvistur, an Icelandic compound word analyzer.
Finally, preliminary experiments show that our model performs very well when evaluated on a closely related language, Faroese.

\section{Compounding in Icelandic}

In Icelandic, any of the open word classes can be combined to form a compound, although some combinations are more productive than others (noun-noun compounding, in particular). The DIM contains a total of 2.9 million unique inflected forms, of which approximately 2.5 million are compounds. Fully split, these compounds are composed of 169,000 distinct compound parts, with 146,000 functioning as heads and 40,000 as modifiers, with some overlap \cite{Dadason_2014}. The figures are for word forms, not lemmas, and the discrepancy in numbers is due to the fact that the heads of compounds are inflected, whereas the modifiers rarely are \cite{Bjarnadottir_2017}.

Any word form can, in theory, appear as a head, and a compound always has the same grammatical features as its head. The form of modifiers is much more restricted, as reflected by the compound part frequencies in the DIM. Nominal modifiers are mostly limited to stems (e.g., \textit{fótbolti} ‘football’, \textit{fótur} ‘foot’ + \textit{bolti} ‘ball’) and inflected forms in the genitive case, singular or plural (e.g., \textit{umferðarskilti} ‘traffic sign’, \textit{umferð} ‘traffic’ + \textit{skilti} ‘sign’). Less commonly, they may appear in the dative case (e.g., \textit{snæviþakinn} ‘snow covered’, \textit{snær} ‘snow’ + \textit{þakinn} ‘covered’), and they can also contain linking elements  (e.g., \textit{hæfnispróf} ‘aptitude test’, \textit{hæfni} ‘competency’ + \textit{próf} ‘test’). Modifiers of other word classes are similarly restricted \cite{Bjarnadottir_2012}.

\section{Related Work}
\newcite{Brown_2002} presents a method for splitting compound words using a bilingual dictionary between a compounding and a non-compounding language. This method identifies cognates, such as ``Abdominalangiographie'' in German and ``abdominal angiography'' in English, finding the split that maximizes the similarity between the constituent parts of the compound and the words in the non-compound.

% The author evaluates the method on a sample of 500 manually split German words using different parameters and reports an accuracy ranging from 99.4\% with a recall of 0.5\% to 82.6\% at a recall of 34.4\%.

\newcite{Koehn_2003} find all ways that a potential compound word can be split into a sequence of known words, optionally joined by linking elements and inflectional suffixes. Each split is scored by the geometric mean of the word frequencies of its constituent parts. Additionally, PoS tags and a translation dictionary are used to filter out implausible splits. When evaluated on a collection of 3,498 German words that have been manually split, this method achieves a precision of 93.8\% and a recall of 90.1\%.

\newcite{Schiller_2005} uses a morphological analyzer to find all possible splits for a given word. Each constituent part is weighted by its frequency in a morphologically annotated corpus. The probability of each potential split is calculated as the product of those weights. This method achieves a precision of 97.7\% and a recall of 99.1\%, when evaluated on a corpus of 30,891 compounds from German news articles.

\newcite{Riedl_2016} propose an unsupervised method for compound splitting based on distributional semantics. Their method finds the split which maximizes the semantic similarity between the compound parts and the compound itself. The method achieves a precision of 96.1\% and a recall of 88.1\%, when evaluated on a set of 158,653 German nouns from newspaper articles.

\newcite{Tuggener_2018} evaluates various neural network architectures, both supervised and unsupervised, for splitting German compounds. The best performance is achieved by a model composed of an unsupervised BiLSTM network combined with a supervised multilayer perceptron (MLP), which obtains an accuracy of 95.1\% when evaluated on a corpus of 75,000 manually split German compounds.

\subsection{Kvistur 1.0}
\label{sec:kvistur}
Kvistur 1.0 is a compound word analyzer, capable of determining the constituent structure of Icelandic word forms \cite{Dadason_2014}. It is trained on a large corpus of manually split compound words from DIM. From this corpus, Kvistur learns the probability that any two compound parts can be combined to form a compound. The probability of a previously unseen combination of constituent parts is estimated from the probability of the former part appearing as a modifier in the corpus and the second part appearing as a head. Kvistur finds all possible ways to split a given word into a sequence of known constituent parts and uses these probabilities to find the likeliest constituent structure. This approach achieves a precision of 97.6\% and a recall of 98.0\%, when evaluated on a sample of 6,098 words from Icelandic Wikipedia articles.

One shortcoming of this method is that it cannot correctly split a compound word if any of its constituent parts are unknown. Additionally, it does not take any semantic information into account when estimating the probability of two parts being combined.

The aim of the work presented in this paper is to develop a neural network-based method for splitting compound words, to evaluate it, and to compare it to the original approach. This method will be integrated into Kvistur\footnote{Kvistur is available at \url{https://github.com/jonfd/kvistur}}.

\section{Method}
We propose a character-based BiLSTM model for splitting compound words. For every character in an input word, the model predicts whether it marks the beginning of the second part of a compound (i.e., the head). The model returns an output vector of equal length to the word form, as shown in Table \ref{table:example}. 

\begin{table}[!h]
\begin{center}
\begin{tabularx}{\columnwidth}{|X|c|c|c|c|c|c|c|c|c|c|c|c|}
  \hline
  \textbf{Input}  & r & a & f & o & r & k & u & þ & ö & r & f\\
  \hline
  \textbf{Output} & 0 & 0 & 0 & 0 & 0 & 0 & 0 & 1 & 0 & 0 & 0\\
  \hline
\end{tabularx}
\caption{An example of an input word (\textit{raforkuþörf}, ‘electric energy requirement‘) and the expected output vector from the model, denoting where a binary split occurs. The output vector for a base word would consist entirely of zeros.}
\label{table:example}
\end{center}
\end{table}

Each element of the vector is equal to 0, except for the element corresponding to the position where a binary split occurs, in which case it is equal to 1. The full constituent structure of a word form can be derived by repeatedly applying the model until no further splits can be made.

BiLSTM models have been shown to perform extremely well on morphologically complex languages, and are used in current state-of-the-art models for Icelandic PoS tagging \cite{Steingrimsson_2019} and named entity recognition \cite{Ingolfsdottir_2019}. Furthermore, character-level embeddings capture information on the internal structure of words and have proven to be very helpful when dealing with OOV words, improving performance for a variety of NLP tasks \cite{Plank_2016,Dos_2014,Verwimp_2017}.

\begin{figure}[!h]
\begin{center}
\includegraphics[width=\columnwidth]{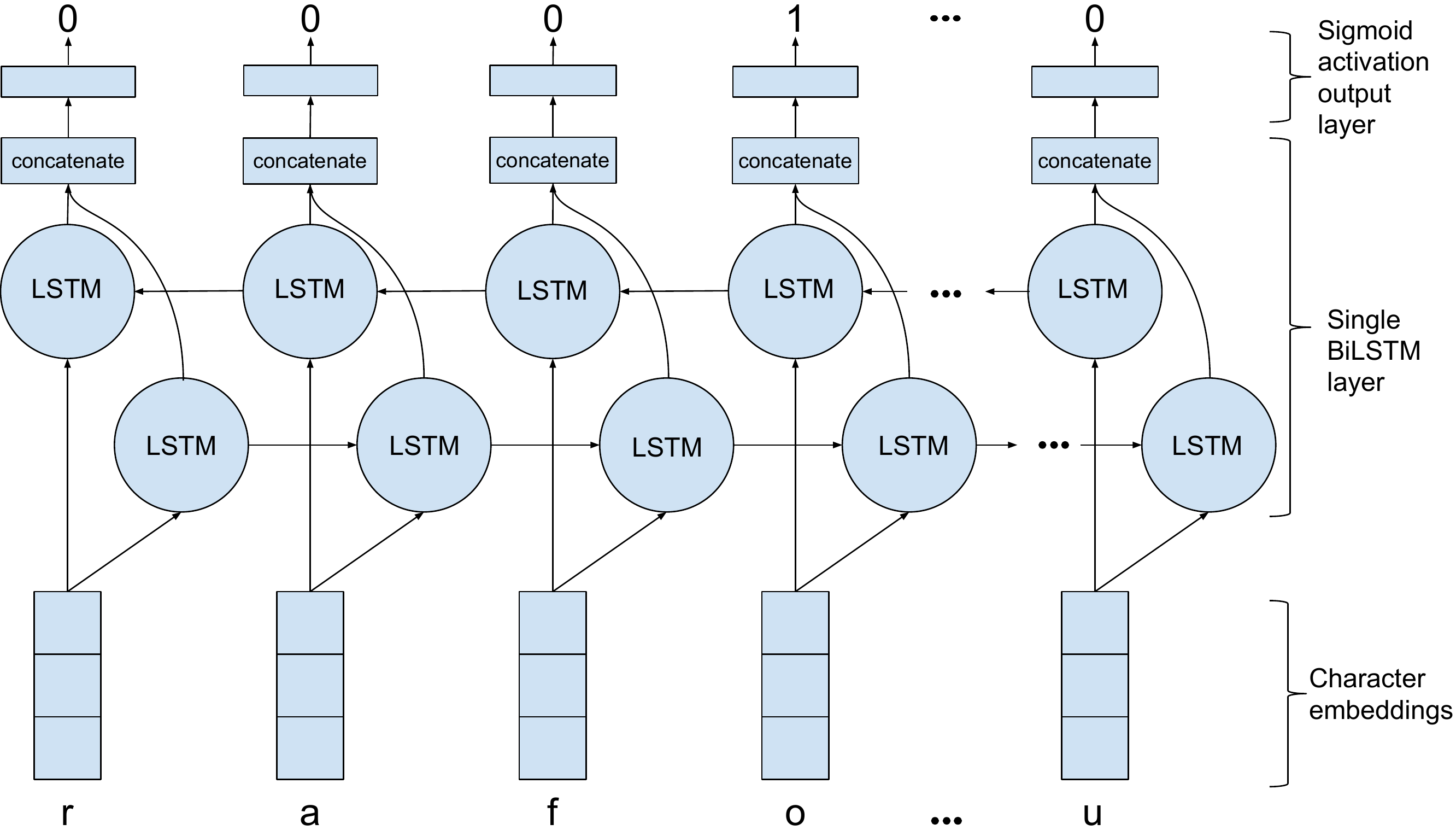}
\caption{A representation of the model with one BiLSTM layer, showing where the compound word \textit{raforku} ‘electric energy’ is split in two.}
\label{fig:model}
\end{center}
\end{figure}

Our model uses character embeddings as input. These embeddings are input into a BiLSTM layer, whose output is fed into a dense output layer, which makes a binary prediction for each character. The model is shown in Figure \ref{fig:model}. We also evaluate a version of this model with two BiLSTM layers.

\newcite{Tuggener_2018} evaluates various neural-network based models for splitting compound words, the best performing of which is an unsupervised BiLSTM model combined with a supervised MLP. The BiLSTM model is used to monitor the probability of an end-of-token symbol appearing after any character within a given token sequence. This probability distribution is passed to a supervised MLP, which is trained to identify the likeliest position of a binary split within a given token. Our model is based on the unsupervised BiLSTM model from this approach. However, our model is fully supervised, directly predicting the likeliest position of a binary split from the training data. With this in mind, we replace the MLP with a simpler sigmoid activation layer.

\section{Experimental Setup}
In this section, we describe the datasets used in our experiments and give an overview of various implementation details.
% and explain how we measure the performance of our model.

\subsection{Datasets}
We evaluate our model on Icelandic and German corpora that have been manually annotated with information on how compound words should be split.

\subsubsection{Icelandic}
For Icelandic, we evaluate our model using data intended for inclusion in \textit{MorphIce} \cite{Bjarnadottir_2019}, a morphological database which will contain the constituent structure of every word form in DIM. In total, this preliminary (unpublished) version of MorphIce contains the constituent structures of approximately 2.9 million unique word forms, of which 2.5 million are compounds. For every unique word form in the data, we create a single training example consisting of the word form as input and a target vector as output, as shown in Table \ref{table:example}.

We observe that there are only approximately 500 word forms (out of 2.9 million) in the dataset with multiple constituent structures. One example is \textit{heimsenda}, which can be split as \textit{heim} ‘home’ + \textit{senda} ‘deliver’ (‘home deliver’) or \textit{heims} ‘world's’ + \textit{enda} ‘end’ (‘apocalypse’). Furthermore, we find that converting all words forms to lowercase does not introduce any additional ambiguity.

We use 80\% of the dataset for training, reserving 10\% for validation and 10\% for testing. Each set consists of a unique set of word forms (i.e., each word form only occurs in one of the three sets), with the exact same proportion of base words and compounds. This means that there is no overlap between any of the three sets, and all words in the validation and test sets are unknown to our model. Additionally, all inflected forms of the same word are placed into the same set.

\subsubsection{German}
For German, we use approximately 75,000 compound words from GermaNet \cite{Henrich_2010}. Each compound has been manually annotated to indicate the position where a binary split occurs. We split the dataset into a training, validation and test set with the same ratios as for Icelandic.

\subsection{Implementation details}
We built our model using Keras\footnote{\url{https://keras.io/}}. We train using the Adam optimizer, with a constant learning rate of 0.001. We train our model for 100 epochs, stopping early if the validation accuracy does not improve for 20 epochs, and select the model with the highest validation accuracy. The character embeddings have 128 dimensions and each BiLSTM layer has 128 hidden units in each direction. The model accepts 40 character long words as input, with shorter words being padded.

\section{Results}
In this section, we compare our proposed model against the statistical-based method used by Kvistur 1.0, which has been shown to achieve high accuracy for splitting Icelandic compounds (see Section \ref{sec:kvistur}).

Table \ref{table:accuracy} summarizes our results on the validation and test data, showing the overall accuracy of the three models. The accuracy is computed as the number of compounds that were correctly split into two parts and base words that were not split, divided by the total number of words. We find that the BiLSTM models significantly outperform the statistical method implemented in Kvistur 1.0, with the two-layer model performing marginally better. Additionally, we find that the two-layer BiLSTM model obtains slightly higher accuracy on German than reported by \newcite{Tuggener_2018}, 96.2\% compared to 95.1\%. We note, however, that although we used the same corpus for training and evaluation, we use a different data split.

\begin{table}[!h]
\begin{center}
\begin{tabularx}{\columnwidth}{|l|>{\centering\arraybackslash}X|>{\centering\arraybackslash}X|}
    \hline
    \textbf{Model} & \textbf{is} & \textbf{de}\\
    \hline
    Kvistur 1.0 & 91.7\% / 91.7\% & -\\
    \hline
    BiLSTM (1 layer) & 97.1\% / 97.1\% & 96.4\% / 95.4\% \\
    \hline
    BiLSTM (2 layers) & 97.4\% / 97.5\% & 96.6\% / 96.2\% \\
    \hline
\end{tabularx}
\caption{The validation and test accuracy (respectively) of the statistical method in Kvistur 1.0 and the BiLSTM models. Kvistur 1.0 was not evaluated on the German dataset as it requires the full constituent structure of compound words for training.}
\label{table:accuracy}
\end{center}
\end{table}

Table \ref{table:base_compound} shows the accuracy of the the three models on base words and compounds in the Icelandic test set. We observe that in addition to a much higher overall accuracy, the BiLSTM models are significantly less likely to mistakenly split base words.

\begin{table}[!h]
\begin{center}
\begin{tabularx}{\columnwidth}{|l|>{\raggedleft\arraybackslash}X|>{\raggedleft\arraybackslash}X|}
    \hline
    \textbf{Model} & \textbf{Base words} & \textbf{Compounds}\\
    \hline
    Kvistur 1.0 & 84.4\% & 92.9\%\\
    \hline
    BiLSTM (1 layer) & 96.2\% & 97.3\% \\
    \hline
    BiLSTM (2 layers) & 96.7\% & 97.6\% \\
    \hline
\end{tabularx}
\caption{The accuracy of the three models with regards to base words and compounds in the Icelandic test set.}
\label{table:base_compound}
\end{center}
\end{table}

We also evaluate the three models in terms of precision and recall, in a similar manner as done by \newcite{Koehn_2003}. The precision is measured as the number of compounds that were correctly split into two parts divided by the total number of words that were split by the model. The recall is measured as the number of compounds that were correctly split into two parts divided by the total number of compounds.

\begin{table}[!h]
\begin{center}
\begin{tabularx}{\columnwidth}{|l|>{\raggedleft\arraybackslash}X|>{\raggedleft\arraybackslash}X|>{\raggedleft\arraybackslash}X|}
    \hline
    \textbf{Model} & \textbf{Precision} & \textbf{Recall} & \textbf{F-score} \\
    \hline
    Kvistur 1.0 & 93.73\% & 92.87\% & 93.30\%\\
    \hline
    BiLSTM (1 layer) & 98.30\% & 97.26\% & 97.78\%\\
    \hline
    BiLSTM (2 layers) & 98.05\% & 97.59\% & 97.82\%\\
    \hline
\end{tabularx}
\caption{The precision, recall and F-score of the three models, as measured on the Icelandic test set.}
\label{table:precision}
\end{center}
\end{table}

The results of the evaluation show that the BiLSTM models obtain a much higher precision and recall than the statistical method in Kvistur 1.0. However, adding more layers to the BiLSTM model does not appear to have a significant impact on precision and recall.

\begin{figure}[!h]
\begin{center}
\includegraphics[width=\columnwidth]{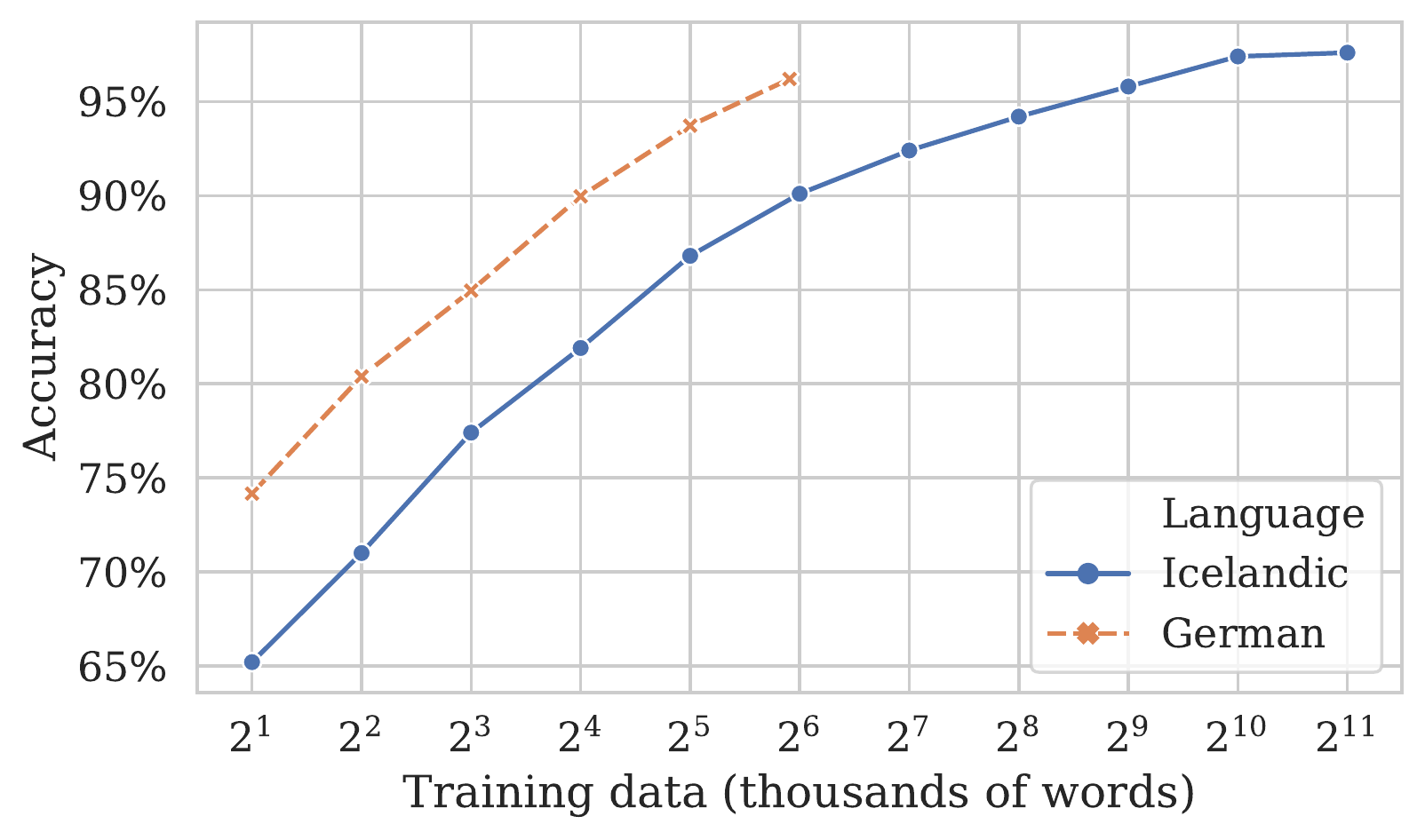}
\caption{Accuracy of the two-layer BiLSTM model for a varying amount of training data.}
\label{fig:training}
\end{center}
\end{figure}

Finally, we evaluate the two-layer BiLSTM model using a varying amount of Icelandic and German training data. We start off by evaluating the accuracy of our model on a training set of 2,000 words, doubling the amount of training data thereafter. We add words to the training data from the complete training set in a decreasing order of frequency, using word frequencies from the Icelandic Gigaword Corpus \cite{Steingrimsson_2018} and from the German Wikipedia\footnote{\url{https://linguatools.org/tools/corpora/wikipedia-monolingual-corpora/}}. The results of this evaluation can be seen in Figure \ref{fig:training}.

Our evaluation shows that increasing the amount of training data has a positive impact on the accuracy of our model up to a certain point. For Icelandic, there are negligible gains from adding more training data beyond two million words. We also observe that our model obtains approximately 10\% higher accuracy for German than for Icelandic, given the same amount of training data. This can perhaps be explained by the more complex morphology of Icelandic.

\section{Conclusion}
In this work, we presented a BiLSTM model for splitting compound words. We evaluated it on manually annotated corpora of Icelandic and German words and found that it outperformed previously evaluated methods.

For future work, we intend to experiment with using constituent part embeddings, derived from MorphIce, in addition to character embeddings. We also want to examine the possibility of applying our model on other languages using transfer learning. We note that without any additional adjustments or training, our model performs very well on Faroese, a low-resource language that is closely related to Icelandic. In a trial run on a sample of 166 Faroese compounds derived from the translations of Icelandic headwords starting with ``dr-'' in ISLEX, an Icelandic-Faroese dictionary, Kvistur 2.0 returned 12 errors: 6 incorrect splits and 6 compounds that were not split. The remaining 154 compounds were correctly split. The sample was hand-checked, but it is too small for evaluation. It should be noted that Faroese spelling differs quite a bit from Icelandic spelling, and so do the rules of compound formation to a degree.

% útskýra "errors" betur. Já og 12 errors: 6 errors er dáldið vont!
% TODO: Footnote með URL eða reference á ISLEX?: >> ISLEX. Þórdís Úlfarsdóttir (ed.). Reykjavík: The Árni Magnússon Institute for Icelandic Studies. <http://islex.is>

% \nocite{*}
\section{Bibliographical References}
\label{main:ref}

\bibliographystyle{lrec}
\bibliography{kvistur}

%\section{Language Resource References}
%\label{lr:ref}
%\bibliographystylelanguageresource{lrec}
%\bibliographylanguageresource{lrec2020W-xample}

\end{document}